\documentclass[conference]{IEEEtran}
\IEEEoverridecommandlockouts
\usepackage{cite}
\usepackage{hyperref}
\usepackage{amsmath,amssymb,amsfonts}
\usepackage{algorithmic}
\usepackage{graphicx}
\usepackage{textcomp}
\usepackage{xcolor}
\usepackage{cleveref}
\usepackage{booktabs}
\usepackage{url} 

\def\BibTeX{{\rm B\kern-.05em{\sc i\kern-.025em b}\kern-.08em
    T\kern-.1667em\lower.7ex\hbox{E}\kern-.125emX}}
\begin{document}

\title{Band-Attention Modulation Network for Robust Face Forgery Detection}


\author{
\IEEEauthorblockN{Zhida Zhang}
\IEEEauthorblockA{\textit{School of Artificial Intelligence, } \\
\textit{UCAS \& NLPR, CASIA} \\
Beijing, China \\
zhida.zhang@cripac.ia.ac.cn
}
\and
\IEEEauthorblockN{Wenkui Yang}
\IEEEauthorblockA{\textit{School of Advanced Interdisciplinary Sciences, } \\
\textit{UCAS \& NLPR, CASIA} \\
Beijing, China \\
wenkui.yang@cripac.ia.ac.cn
}
\and
\IEEEauthorblockN{Xinlei Ma}
\IEEEauthorblockA{\textit{School of Advanced Interdisciplinary Sciences, } \\
\textit{UCAS \& NLPR, CASIA} \\
Beijing, China \\
xinlei.ma@nlpr.ia.ac.cn
}
\and
\IEEEauthorblockN{Qihang Fan}
\IEEEauthorblockA{\textit{School of Artificial Intelligence, } \\
\textit{UCAS \& NLPR, CASIA} \\
Beijing, China \\
fanqihang.159@gmail.com
}
\and
\IEEEauthorblockN{Jie Cao \textsuperscript{*}}
\IEEEauthorblockA{\textit{NLPR, CASIA} \\
Beijing, China \\
jie.cao@cripac.ia.ac.cn
}

\thanks{\textsuperscript{*} is the corresponding author. This work was supported by the Beijing Natural Science Foundation (Grant No. L257008 \& L252145) and National Natural Science Foundation of China (Grant No. 62576338 \& 62550062 \& 32341009).}
}

\maketitle

\begin{abstract}
Face forgery detection faces critical challenges in generalizing to unseen manipulation techniques and remaining robust under image compression, which often obscures subtle artifacts. Existing methods typically rely on fixed filters or coarse band separation, lacking the adaptability to learn task-specific spectral cues. 
To address this, we propose the Band-Attention Modulation Network (BAM-Net), a novel framework that pioneers learnable, fine-grained modulation of frequency components for forgery detection. At its core is the Band-Attention Modulation (BAM) mechanism, which transforms an image into its Discrete Cosine Transform (DCT) spectrogram and learns to dynamically reweight frequency bands along anti-diagonals. 
This process effectively enhances forgery-related spectral signatures while suppressing less informative ones, simulating an adaptive “inverse compression” that counters information loss. The modulated frequency information is then fused with the spatial domain to guide a lightweight yet effective spatial backbone equipped with distance-decayed attention for comprehensive feature extraction. 
Extensive experiments on FaceForensics++, Celeb-DF, and DFDC datasets demonstrate that BAM-Net achieves state-of-the-art performance. More importantly, it exhibits exceptional generalization in cross-dataset, cross-compression, and cross-manipulation scenarios, underscoring the vital role of adaptive frequency band modulation in building robust forgery detectors.
\end{abstract}

\begin{IEEEkeywords}
Face Forgery Detection, Frequency Domain, Band Modulation, Adaptive Filtering, Generalization
\end{IEEEkeywords}

\section{Introduction}
\label{sec:intro}

\begin{figure*}[htbp]
  \centering
  \includegraphics[width = 0.75\textwidth]{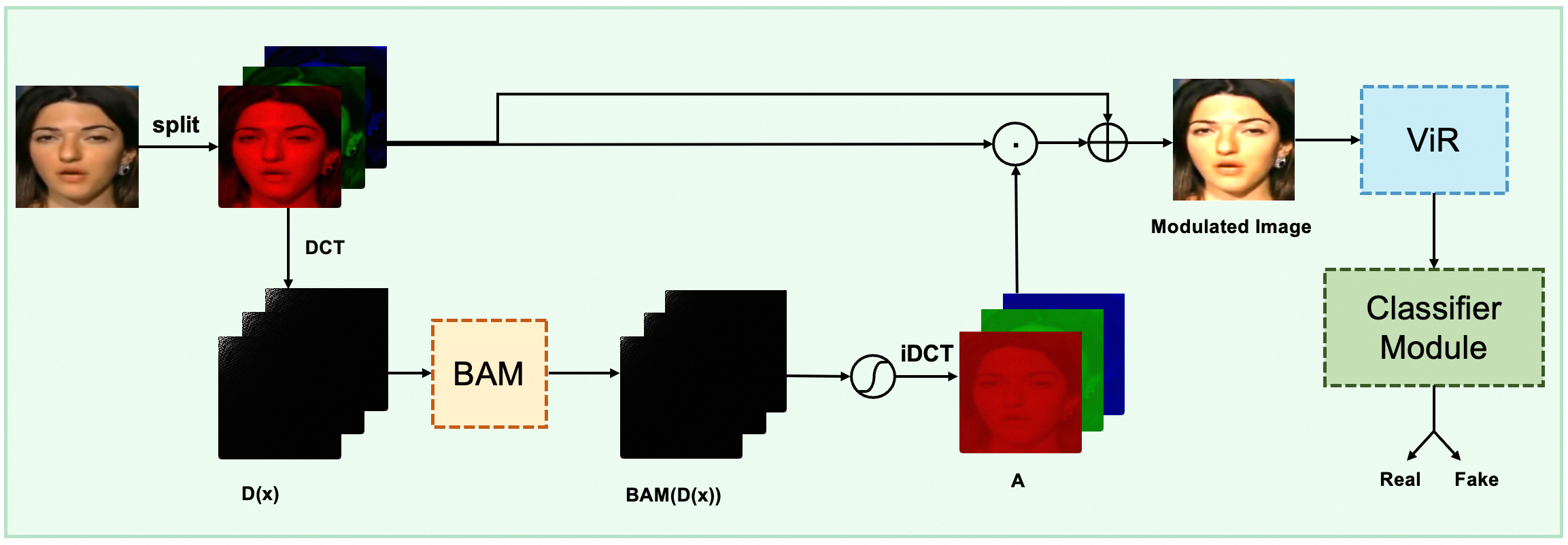}
  \caption{Overview of the proposed Band-Attention Modulation Network (BAM-Net). 
  }
  \label{fig:overview}
\end{figure*}

The emergence of highly realistic synthetic face media, driven by advances in generative AI, poses significant threats to digital trust and security. Detecting such forgeries has thus become an urgent and challenging task in computer vision and multimedia forensics.

Current detection methodologies can be broadly divided into two categories. The first operates primarily in the \textbf{spatial domain}, utilizing convolutional neural networks (CNNs) or vision transformers to capture local texture inconsistencies or global semantic anomalies \cite{rossler2019faceforensics++, F2Trans, duan2025dual}. While effective under controlled conditions, these approaches often suffer from performance degradation when faced with strong image compression (e.g., JPEG) or novel forgery techniques not seen during training, limiting their practical applicability. The second category explores the \textbf{frequency domain}, motivated by the observation that forgery operations introduce distinctive spectral distortions \cite{xu2020learning, F3Net}. However, most existing frequency-based methods employ hand-crafted high-pass filters or static band separation, which are suboptimal as they cannot adapt to the diverse and evolving spectral patterns of different forgery methods and compression artifacts.

We argue that the key to a robust and generalizable detector lies in \textit{adaptively emphasizing the most discriminative frequency components} that are indicative of forgeries, while effectively integrating this spectral insight with spatial context analysis. To this end, we introduce the \textbf{Band-Attention Modulation Network (BAM-Net)}. The cornerstone of BAM-Net is a novel \textbf{Band-Attention Modulation (BAM)} mechanism. Unlike prior static approaches, BAM performs \textit{learnable, fine-grained reweighting of frequency bands}. Specifically, it decomposes an image's DCT spectrogram into frequency bands along anti-diagonals and employs a lightweight network to predict a dynamic weight for each band. This allows the model to automatically amplify frequencies crucial for detecting forgeries (often attenuated by compression) and attenuate less relevant ones, effectively implementing a data-driven “spectral enhancer”.

To construct a complete and efficient detector, the modulated spectral information needs to be seamlessly incorporated into spatial feature learning. To achieve this, we require a spatial backbone that is both effective in modeling long-range context and computationally efficient. Drawing inspiration from the decay mechanism in Retentive Networks (RetNet) \cite{RetNet} designed for sequence modeling, we develop a \textbf{Vision Retentive (ViR)} module tailored for image data. Crucially, ViR is \textit{adapted and extended} for our 2D visual domain: it utilizes a \textbf{distance-dependent spatial decay matrix} based on Manhattan distance to incorporate explicit 2D spatial priors. This design inherently favors attention to local regions where fine-grained forgery artifacts often reside, while still maintaining a global receptive field. Furthermore, we employ an axial decomposition strategy to reduce the computational complexity to linear, ensuring efficiency especially in early layers with a large number of tokens. The ViR module thus serves as a powerful yet efficient spatial context encoder that works in concert with the front-end BAM module within our \textbf{BAM-Net} framework.

As illustrated in \cref{fig:overview}, BAM-Net integrates the BAM front-end with the spatial backbone into a cohesive pipeline: the BAM module first performs adaptive frequency band modulation, whose output is then projected to the spatial domain to guide the subsequent spatial feature extraction. This synergistic design enables the model to leverage complementary forgery cues from both spectral and spatial domains.

We conduct comprehensive evaluations on standard benchmarks including FaceForensics++ (FF++), Celeb-DF, and DFDC. BAM-Net not only achieves state-of-the-art performance in intra-dataset settings but, more significantly, demonstrates superior generalization capability in challenging cross-dataset, cross-compression, and cross-manipulation tests. Ablation studies confirm that the proposed BAM mechanism is the primary contributor to this robust generalization performance.

Our contributions are threefold:
\begin{itemize}
    \item We propose a novel \textbf{Band Modulation (BAM)} mechanism, which introduces \textit{learnable, dynamic frequency band reweighting} to face forgery detection. This represents a shift from static spectral filtering to adaptive spectral modulation, significantly enhancing model robustness and generalization.
    \item We develop an efficient \textbf{spatial backbone} with distance-decayed attention and 2D spatial priors, providing effective global context modeling with linear complexity, which serves as a strong complementary component to our frequency modulation front-end.
    \item By integrating BAM with the spatial backbone, we present \textbf{BAM-Net}, a holistic framework that achieves state-of-the-art performance. BAM-Net establishes a new paradigm for improving forgery detection generalization through adaptive frequency domain modulation.
\end{itemize}

\section{Method}

Our proposed Band-Attention Modulation Network (BAM-Net) is designed to detect face forgeries by adaptively emphasizing discriminative frequency cues and efficiently modeling spatial context. As shown in \cref{fig:overview}, BAM-Net consists of two components: (1) the \textbf{BAM} front-end, which reweights frequency bands to highlight forgery artifacts, and (2) the \textbf{ViR} backbone, which models long-range dependencies with explicit 2D priors and linear complexity. In this section, we first detail the BAM mechanism (Section \cref{sec:bam}), then describe the ViR backbone (Section \cref{sec:vir}), and finally present the integrated pipeline (Section \cref{sec:pipeline}).

\subsection{Band-Attention Modulation (BAM)}
\label{sec:bam}

\begin{figure}[tbp]
  \centering
  \includegraphics[width=0.48\textwidth]{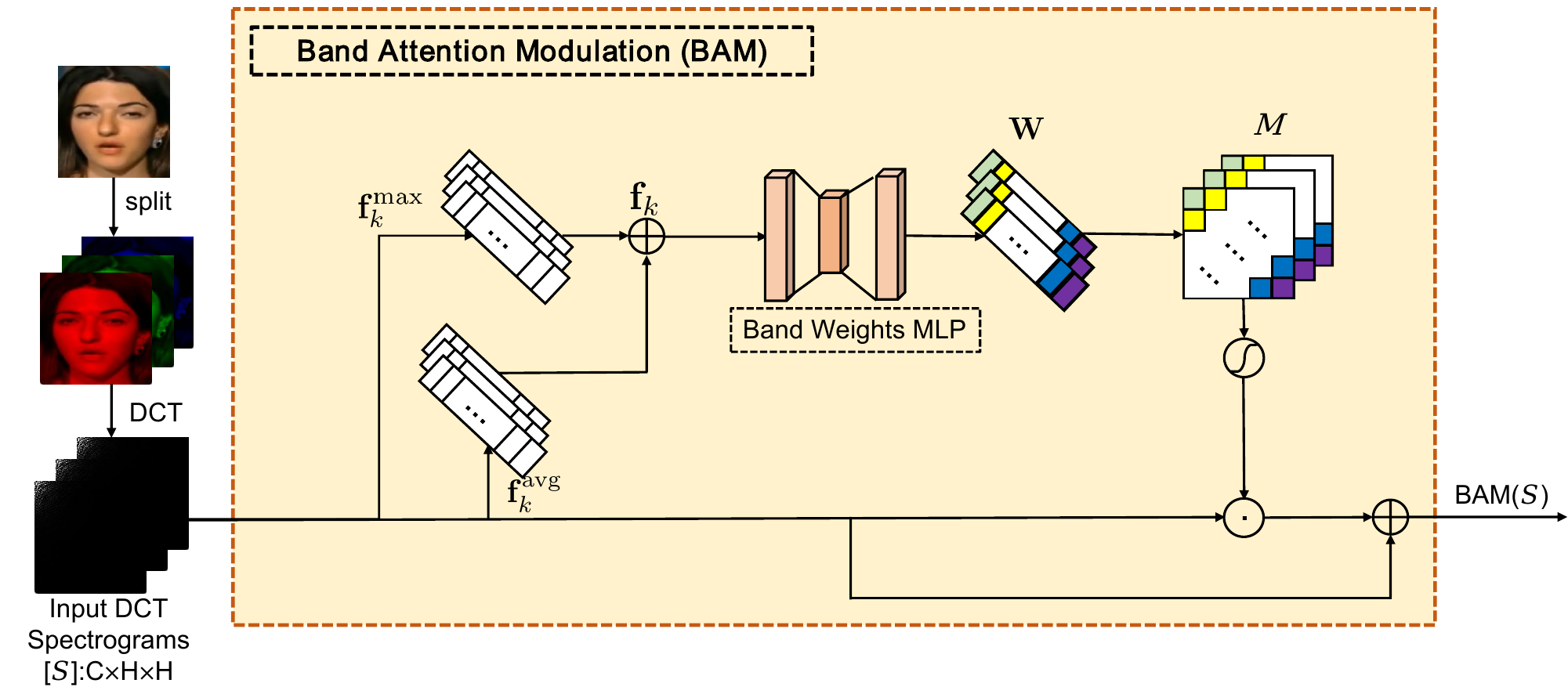}
  \caption{Detailed diagram of the Band-Attention Modulation (BAM) mechanism. 
  Input image is transformed via DCT. Coefficients on the same anti-diagonal are treated as a frequency band. Band-level features are extracted via pooling, processed by an MLP to produce band weights, which are reshaped into a 2D attention map to modulate the original spectrogram.
  }
  \label{fig:bam_detail}
  \vspace{-1em}
\end{figure}

The key to robust forgery detection lies in capturing subtle artifacts that are often embedded in specific frequency components and can be obscured by image compression. To move beyond static, hand-crafted frequency filters, we propose the \textbf{Band-Attention Modulation (BAM)} mechanism. BAM learns to dynamically reweight different frequency bands, acting as a data-driven ``spectral enhancer'' that emphasizes forgery-relevant frequencies.

\subsubsection{Frequency Band Decomposition via DCT}
Given an input RGB image $x \in \mathbb{R}^{C \times H \times W}$ (with $H=W$ for simplicity), we first apply the 2D Discrete Cosine Transform (DCT) to each channel independently, obtaining the DCT spectrogram $S = \mathcal{D}(x) \in \mathbb{R}^{C \times H \times H}$.
In the spectrogram, we consider coefficients along the same \textbf{anti-diagonal} (from bottom-left to top-right) as belonging to the same frequency band, since they share the same sum of horizontal and vertical frequency indices. This yields $N = 2H - 1$ distinct frequency bands. This decomposition is illustrated in the left part of \cref{fig:bam_detail}.

\subsubsection{Band-wise Feature Extraction}
To characterize each band, we extract two complementary statistics. For the $k$-th anti-diagonal (where $k = 1, \ldots, N$), let $D_k \in \mathbb{R}^{C \times L_k}$ denote the matrix of coefficients along that diagonal, with $L_k$ being its length (which varies with $k$). We compute both the \textbf{maximum} and \textbf{average} value across the spatial dimension for each channel:
\begin{align}
    \mathbf{f}^{\text{max}}_k &= \max_{j} D_{k,:,j}, \quad
    \mathbf{f}^{\text{avg}}_k = \frac{1}{L_k} \sum_j D_{k,:,j},
\end{align}
where both $\mathbf{f}^{\text{max}}_k$ and $\mathbf{f}^{\text{avg}}_k$ are $C$-dimensional vectors. We then sum them to form a compact band descriptor:
\begin{align}
    \mathbf{f}_k = \mathbf{f}^{\text{max}}_k + \mathbf{f}^{\text{avg}}_k \in \mathbb{R}^C.
\end{align}
Stacking these descriptors across all bands gives a band-feature matrix $\mathbf{F} = [\mathbf{f}_1, \ldots, \mathbf{f}_N]^\top \in \mathbb{R}^{N \times C}$.

\subsubsection{Cross-Channel Band-Weight Learning}
The core of BAM is a lightweight neural network that learns to assign an importance weight to each frequency band. We reshape $\mathbf{F}$ to $\mathbf{F}' \in \mathbb{R}^{(B \cdot C) \times N}$, where $B$ is the batch size. This allows the network to \textbf{share parameters across channels and samples} while still producing sample- and channel-specific weights. $\mathbf{F}'$ is processed by a two-layer MLP with batch normalization and GELU activation:
\begin{align}
    \mathbf{W}' &= \mathrm{MLP}(\mathbf{F}') \in \mathbb{R}^{(B \cdot C) \times N}, \\
    \text{where } \mathrm{MLP}(z) &= \mathrm{FC}_2(\mathrm{GELU}(\mathrm{BN}(\mathrm{FC}_1(z)))).
\end{align}
The output is reshaped back to $\mathbf{W} \in \mathbb{R}^{B \times C \times N}$, where each element $w_{b,c,k}$ represents the learned weight for the $k$-th frequency band of the $c$-th channel in the $b$-th sample.

\subsubsection{Spectral Modulation and Reconstruction}
The band weights $\mathbf{W}$ are expanded into a full 2D \textit{attention map} $M \in \mathbb{R}^{B \times C \times H \times H}$ by assigning the weight $w_{b,c,k}$ to all positions on the $k$-th anti-diagonal. The original spectrogram $S$ is then modulated via a residual connection:
\begin{align}
    \tilde{S} = S \odot \sigma(M) + S,
    \label{eq:bam_output}
\end{align}
where $\sigma$ is the Sigmoid function ensuring weights lie in $(0,1)$, and $\odot$ denotes element-wise multiplication. The modulated spectrogram $\tilde{S}$ is transformed back to the spatial domain via the inverse DCT (iDCT), producing an enhanced image that accentuates frequency components deemed important by the learned band attention.

\subsection{Vision Retentive Backbone for Spatial Context}
\label{sec:vir}

To process the modulated image from the BAM front-end and extract comprehensive spatial features, we require an efficient backbone capable of modeling long-range dependencies across facial regions. For this purpose, we design a Vision Retentive (ViR) module. Drawing inspiration from the retention mechanism in Retentive Networks (RetNet) \cite{RetNet} designed for efficient sequence modeling, ViR adapts and extends the core idea of distance-decayed attention to the 2D visual domain. Specifically, ViR introduces a self-attention variant that incorporates explicit 2D spatial priors via a distance-dependent decay bias and rotary position embeddings, encouraging the model to focus more on local regions where fine-grained forgery artifacts often reside while maintaining a global receptive field. This design is particularly tailored for the efficiency requirements of processing high-resolution images.

\subsubsection{Distance-Decayed Attention with Rotary Positions}
Given an input feature map $X \in \mathbb{R}^{B \times H \times W \times C}$, we first project it into queries, keys, and values using learnable linear layers: $Q = XW_Q$, $K = XW_K$, $V = XW_V$. To incorporate relative position information, we apply rotary position embeddings (RoPE) \cite{su2021roformer} to $Q$ and $K$, yielding $Q_r$ and $K_r$.

The core innovation is the introduction of a 2D spatial decay bias. For two tokens at positions $\mathbf{p}_i = (x_i, y_i)$ and $\mathbf{p}_j = (x_j, y_j)$, the decay bias is a function of their Manhattan distance:
\begin{align}
    B_{ij} = (|x_i - x_j| + |y_i - y_j|) \cdot \log \gamma,
    \label{eq:2d_bias}
\end{align}
where $\gamma \in (0,1)$ is a decay factor. In practice, we allow each attention head $h$ to have a different decay rate $\gamma_h$, following an exponential schedule: $\gamma_h = 1 - 2^{-(\alpha + h \cdot \beta)}$, where $\alpha$ and $\beta$ are hyperparameters controlling the initial decay and the decay range across heads. This multi-scale decay enables different heads to focus on different receptive fields.

The attention logits are computed as $A = Q_r K_r^\top / \sqrt{d_k}$, where $d_k$ is the key dimension. The decay bias $B$ is added to $A$ before the Softmax normalization, yielding our \textbf{Retention Self-Attention (ReSA)}:
\begin{align}
    \mathrm{ReSA}(X) = \mathrm{Softmax}\left( \frac{Q_r K_r^\top}{\sqrt{d_k}} + B \right) V.
    \label{eq:resa_full}
\end{align}
The addition of $B$ penalizes attention between distant tokens, explicitly encoding a spatial prior that favors local interactions.

\subsubsection{Axial Decomposition for Linear Complexity}
Direct computation of ReSA over all $HW$ tokens has quadratic complexity $O((HW)^2)$. To achieve linear scaling, we employ an axial decomposition strategy, decomposing the 2D attention into separate horizontal and vertical 1D attentions. This decomposed form, termed \textbf{ReSA-A}, proceeds in two steps:

\textbf{Step 1: Width-wise Attention.} We compute attention along the width dimension for each row independently. For the $y$-th row, the decay bias is $B^W_{ij} = |x_i - x_j| \cdot \log \gamma$. The width-wise attention output $V^W$ is:
\begin{equation}
    \begin{aligned}
    A^W_y = \mathrm{Softmax}\left( \frac{Q_r(y,:) K_r(y,:)^\top}{\sqrt{d_k}} + B^W \right), \\
    V^W(y,:) = A^W_y V(y,:),
    \end{aligned}
\end{equation}
\textbf{Step 2: Height-wise Attention.} We then compute attention along the height dimension for each column of $V^W$. For the $x$-th column, the decay bias is $B^H_{ij} = |y_i - y_j| \cdot \log \gamma$. The final output is:
\begin{equation}
    \begin{aligned}
    A^H_x = \mathrm{Softmax}\left( \frac{Q_r(:,x) K_r(:,x)^\top}{\sqrt{d_k}} + B^H \right), \\
    \mathrm{ReSA\text{-}A}(X)(:,x) = A^H_x V^W(:,x),
    \end{aligned}
\end{equation}

This sequential computation reduces the complexity to $O(HW(H+W))$, which is linear in the number of tokens when $H$ and $W$ are of the same order. Crucially, ReSA-A maintains an effective receptive field equivalent to the full 2D attention, as each token can indirectly attend to all others through the two passes.

\subsubsection{Local Enhancement and Network Architecture}
Following the attention block, we incorporate a local feature enhancement module. Specifically, we apply a depthwise separable convolution (DWConv) with a $5\times5$ kernel to the value projection $V$ before adding it to the attention output:
\begin{align}
    \text{Output} = \mathrm{ReSA}(X) + \mathrm{DWConv}(V).
\end{align}
This ``Locally Enhanced Positional Encoding'' (LePE) strengthens the model's ability to capture fine-grained local patterns, complementing the long-range modeling of attention.

\begin{figure}[t]
  \centering
  \hfill
  \begin{minipage}{0.48\textwidth}
    \centering
    \includegraphics[width=\linewidth]{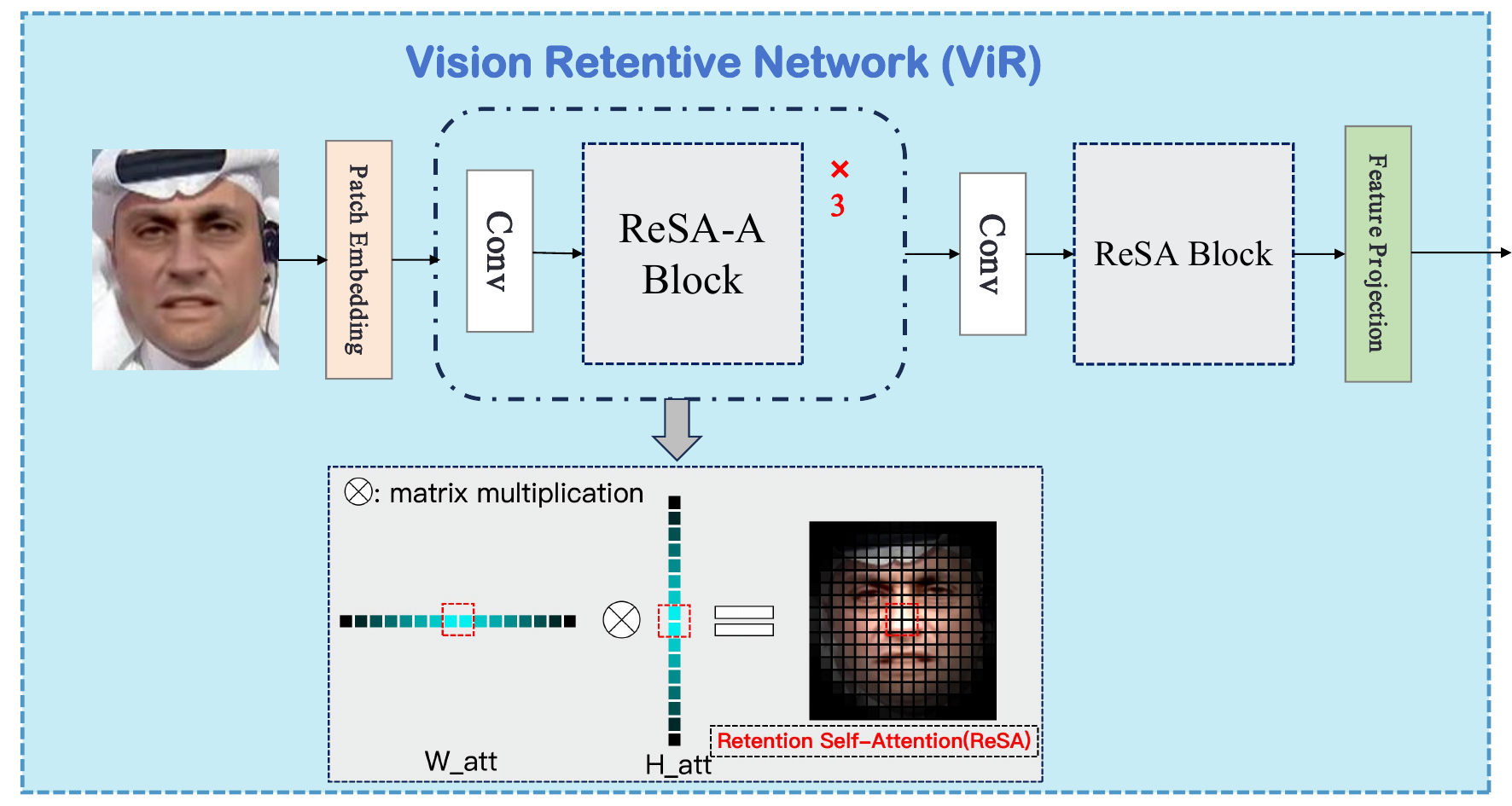}
    \caption{Architecture of the Vision Retentive (ViR) backbone. 
    It consists of four stages. Early stages use linearly complex ReSA-A blocks; the last stage uses full ReSA blocks.
    }
    \label{fig:vir_arch}
  \end{minipage}
  \vspace{-1em}
\end{figure}

As illustrated in \cref{fig:vir_arch}, the ViR backbone adopts a hierarchical four-stage architecture, similar to modern vision transformers \cite{ViT}. Each stage consists of several RetNet blocks. The first three stages employ the efficient ReSA-A blocks to handle the large number of tokens in high-resolution feature maps, while the final stage uses the full ReSA blocks to capture more global interactions at lower spatial resolution. Between stages, a patch merging layer (implemented as a strided convolution) reduces the spatial dimensions and increases the channel count. This design ensures an effective trade-off between computational efficiency and representational power, making ViR a suitable spatial encoder for our forgery detection pipeline.

\subsection{Integrated Pipeline: BAM-Net}
\label{sec:pipeline}

The complete Band-Attention Modulation Network (BAM-Net) seamlessly integrates the BAM front-end with the ViR backbone into a cohesive and efficient pipeline for face forgery detection, as illustrated in \cref{fig:overview}. The process transforms an input image into a forgery probability through the following stages.

\subsubsection{Frequency-Aware Input Modulation}
Given an input RGB image $x \in \mathbb{R}^{3 \times H \times W}$, the first step is adaptive frequency enhancement via the BAM module. As detailed in Section \ref{sec:bam}, we compute its DCT spectrogram $S = \mathcal{D}(x)$, apply band-attention modulation to obtain $\tilde{S} = \mathrm{BAM}(S)$, and then project the modulated spectrogram back to the spatial domain via the inverse DCT. To ensure stable gradients and interpretable modulation, we pass the result through a Sigmoid function, yielding a spatial attention map $A \in [0,1]^{3 \times H \times H}$:
\begin{align}
    A = \sigma \left( \mathcal{D}^{-1} \left( \mathrm{BAM}(\mathcal{D}(x)) \right) \right),
    \label{eq:attmap_gen}
\end{align}
where $\sigma$ denotes the Sigmoid function, and $\mathcal{D}^{-1}$ is the inverse DCT. This attention map $A$ highlights spatial regions where the frequency components deemed discriminative by BAM are concentrated.

The original image $x$ is then element-wise modulated by this attention map via a residual connection, producing an \textbf{enhanced image} $x'$:
\begin{align}
    x' = A \odot x + x.
    \label{eq:spatial_mod}
\end{align}
This operation adaptively amplifies the visual salience of potential forgery traces in the input space, guided by the learned frequency-band importance.

\subsubsection{Spatial Feature Extraction and Classification}
The enhanced image $x'$ serves as the input to the Vision Retentive (ViR) backbone described in Section \ref{sec:vir}. The hierarchical ViR encoder processes $x'$ through its four stages, progressively extracting multi-scale spatial features while efficiently modeling both local and global context through its distance-decayed attention mechanism.

The final feature map from ViR is globally average-pooled to obtain a compact representation $\mathbf{z} \in \mathbb{R}^D$. This representation is then fed into a lightweight classification head, consisting of a dropout layer for regularization followed by a linear projection, to produce the final real/fake prediction $\hat{y}$:
\begin{align}
    \hat{y} = \mathrm{Classifier}(\mathbf{z}) = \mathrm{Linear}(\mathrm{Dropout}(\mathbf{z})).
\end{align}

\begin{table*}[tbp]
\caption{Comparison against state-of-the-art methods on intra-dataset benchmarks (FF++). 
}
\centering
\label{tab:Intra}
\begin{tabular}{@{}lcccccccccccc@{}}
\toprule
Method & \multicolumn{2}{c}{FF++c40(IL)} & \multicolumn{2}{c}{FF++c23(IL)} &  \multicolumn{2}{c}{FF++c40(IVL)}& \multicolumn{2}{c}{FF++c23(IVL)}& \multicolumn{2}{c}{Celeb-DF(IVL)} & \multicolumn{2}{c}{DFDC(IVL)}\\
\cmidrule(lr){2-3} \cmidrule(lr){4-5} \cmidrule(lr){6-7} \cmidrule(lr){8-9} \cmidrule(lr){10-11} \cmidrule(lr){12-13}
& AUC & Acc & AUC & Acc  & AUC & Acc & AUC & Acc  & AUC & Acc & AUC &Acc  \\
\midrule
Xception 
& 81.76 & 80.32 & 94.86 & 92.39 & 89.30& 86.86& 94.80& 93.86& 99.73& 97.9& -&-\\
SPSL 
& 82.82 & 81.57 & 95.32 & 91.50 & -& -& -& -& -& -& -&-\\
$F^3$-Net 
& 87.22 & 84.53 & 97.80 & 93.12 & 93.30& 90.43& 98.10& 97.52& 98.93& 95.95& -&-\\
M2TR 
& 87.97 & 85.09 & 97.84 & 93.22 & 92.46& 88.71& 99.19& 96.14& -& -& -&-\\
LiSiam 
& \underline{91.44} & \underline{87.81} & \underline{99.13} & \underline{96.51} & 94.65& 91.29& 99.52& 97.57& -& -& -&-\\
SIA 
& -& -& -& -& 93.45& 90.23& 99.35& 97.64& \underline{99.94}& 98.59& 91.33&81.20\\
DisGRl 
& -& -& -& -& \underline{95.19}& \underline{91.27}& 99.48& \underline{97.69}& 99.91& \underline{98.71}& \underline{92.50}& \underline{82.35}\\
DPBoT 
& -& -& -& -& 85.64& 82.85&97.56 & 95.35& -& -& -& -\\
DDIN(+PUTT) 
&-&-& -& -& 95.16& 91.23& \underline{99.59}& 97.91 -& -& -& -& -\\
Ours 
& \bf 93.19 & \bf 89.63 & \bf 99.32 & \bf 97.02 & \bf 96.88& \bf 91.86& \bf 99.81& \bf 98.00& \bf 100& \bf 99.39& \bf 94.49& \bf 85.85\\
\bottomrule
\end{tabular}
\vspace{-2em}
\end{table*}

\subsubsection{Summary of Design Rationale}
The BAM-Net pipeline embodies a clear division of labor: the \textbf{BAM module} operates as a \textbf{frequency-domain pre-processor}, learning to highlight subtle spectral artifacts that are often hallmarks of manipulation and are robust to compression. The \textbf{ViR backbone} then acts as a powerful \textbf{spatial-domain analyzer}, capturing the contextual relationships and spatial inconsistencies of the enhanced visual signals. This synergistic, two-stage design allows the model to exploit complementary forgery cues from both the frequency and spatial domains, leading to a detector that is not only accurate but also exhibits strong generalization across different datasets, compression levels, and unseen manipulation techniques—as will be demonstrated in the experiments.

\section{Experiment}

\subsection{Datasets and Evaluation Metrics}
We evaluate BAM-Net on widely used face forgery detection benchmarks: FaceForensics++ (FF++) \cite{rossler2019faceforensics++} (both high-quality `c23' and low-quality `c40' versions), Celeb-DF \cite{li2020celeb}, and DFDC \cite{DFDC}. We report both Image-Level (IL) and Image-based Video-Level (IVL) performance using Area Under the ROC Curve (AUC) and Accuracy (Acc). For IVL, the video score is obtained by averaging the prediction scores of all its frames. 
Detailed dataset partitions, preprocessing steps, and implementation details are provided in the supplementary material.

\subsection{Main Results}

\subsubsection{Intra-Dataset Evaluation}
We first evaluate BAM-Net under the intra-dataset setting on FF++. As shown in Table \ref{tab:Intra}, it achieves top-performing results across both quality levels. On the more challenging FF++c40 subset, BAM-Net attains \textbf{93.19\%} AUC and \textbf{89.63\%} image-level accuracy, showing a clear advantage under compression. On the high-quality FF++c23 subset, it also reaches the highest AUC (\textbf{99.32\%}) and accuracy (\textbf{97.02\%}). The consistent superiority in video-level evaluations further confirms the robustness of our per-frame predictions.
\subsubsection{Cross-Dataset Evaluation}
Generalization to unseen datasets is a critical challenge. Table \ref{tab:corss_dataset} presents cross-dataset results where models are trained on FF++ and tested on Celeb-DF and DFDC. When trained on FF++c40, BAM-Net yields the best image-level AUC on both target datasets. At the video level, it delivers highly competitive performance, closely matching or surpassing recent methods without leveraging any temporal modeling. Training on higher-quality FF++c23 further improves generalization, with BAM-Net achieving leading or comparable results to state-of-the-art methods, including video-based FTCN \cite{FTCN}. This indicates that our image-based model captures forgery cues that are highly transferable across different data distributions.

\begin{table}[tbp]
\caption{Cross-dataset evaluation of the FF++ dataset by AUC (\%). 
}
\centering
\setlength{\tabcolsep}{3pt} 
\label{tab:corss_dataset}
\begin{tabular}{lccc|cc}
\hline
Methods & Training & \multicolumn{2}{c|}{Testing(IL)}  & \multicolumn{2}{c}{Testing(IVL)} \\
& & Celeb & DFDC & Celeb & DFDC \\
\hline
Xception 
& & 65.50 & 69.70 & - & - \\
M2TR 
& & 66.15 &70.29 & 72.05 & 66.02 \\
$F^3$-Net 
& FF++ & 71.29 & 65.39 & 77.92 & 67.35 \\
Two-branch 
& (c40) & 73.41 & - & 76.70 & -  \\
DeepFake-Adapter 
& & - & - & 71.74 & 72.66  \\
DPBoT 
& & - & - & 78.15 & \bf 74.39  \\
FDPNet 
& & - & - & \bf 82.46 & 74.03 \\
DDIN(+PUTT) 
& & - & - & 69.68 & 71.31  \\
\bf Ours & & \bf78.74 & \bf 72.42 & \underline{82.44} & \underline{74.28}  \\
\hline
$F^3$-Net 
& & 68.69 & 67.45 & 75.68 & 70.88 \\
Xception 
& & 65.30 & 72.20 & 73.70 & 70.90\\
MADD 
& & 67.44 & - & 76.65 & 67.34\\
M2TR 
& FF++ & 69.94 & 69.94 & 75.76& 73.12 \\
Face X-ray 
& (c23) & - & - & 79.50 & 65.50 \\
LTW 
& & - & - & 77.14 & \underline{74.58} \\
FTCN 
& & - & - & \bf86.90 & 74.00 \\

MTSD 
& & \underline{75.90} & \bf75.90 & - & - \\
\bf Ours & & \bf81.85 & \underline{72.37} & \underline{85.85}& \bf 75.51 \\
\bottomrule
\vspace{-2em}
\end{tabular}
\end{table}

\subsubsection{Cross-Manipulation Evaluation}
To assess robustness against specific unseen forgery techniques, we follow the cross-manipulation protocol on FF++c40: training on three manipulation methods and testing on the held-out one. Results in Table \ref{tab:cross_mani:3to1} show that BAM-Net outperforms all compared approaches in both the GID-DF and GID-FF settings. On GID-DF, we achieve an AUC of \textbf{89.91\%}, surpassing the previous best method by a noticeable margin. This demonstrates BAM-Net's strong capability to learn manipulation-agnostic forensic features.

\subsection{Generalization Discussion}
BAM-Net shows better generalization in both cross-dataset and cross-manipulation evaluations. We attribute this to two key components: (1) the \textbf{BAM front-end}, which adaptively modulates frequency bands. By emphasizing or suppressing informative spectral components, it enhances features that are less sensitive to compression and more consistent across generation sources. (2) the \textbf{ViR backbone}, whose distance-decayed attention prioritizes local interactions while maintaining global context. This inductive bias helps the model focus on intrinsic local inconsistencies shared across manipulation methods. Together, frequency-domain modulation and spatially aware feature extraction provide a strong prior for generalizable deepfake detection.

\begin{table}[tbp]
\centering
\caption{Cross-manipulation evaluation on FF++c40. 
}
\begin{tabular}{l|c c|c c}
\hline
Methods & \multicolumn{2}{c|}{GID-DF} & \multicolumn{2}{c|}{GID-FF} \\
\cline{2-5}
& AUC & Acc & AUC & Acc \\
\hline
EfficientNet 
& 75.30& 67.60& 67.40& 61.41\\
ForensicTransfer 
& - & 68.20& - & 55.00 \\
Multi-task 
& - & 66.76& - & 56.50 \\
MLDG 
& 73.12& 67.15& 61.70& 58.12\\
LTW 
& 75.60 & 69.15 & 72.40& 65.70\\
DCL 
& 83.82& 75.90& 75.07& 67.85\\
M2TR 
& 84.85& 74.29& 71.70& 66.43\\
$F^3$ Net 
& 85.77& 77.50& 73.70& 64.64\\
DDIN(PUTT) 
& 87.57& 79.35& 79.83& 69.33\\
\bf Ours
& \bf 89.91 & \bf 82.86 & \bf 83.23& \bf 73.93 \\
\hline
\end{tabular}
\vspace{-1em}
\label{tab:cross_mani:3to1}
\end{table}

\subsection{Ablation Study}

\subsubsection{Impact of BAM}
Table \ref{tab:eval_all} compares ViR with and without the BAM module. While ViR alone performs well on intra-dataset tests, adding BAM brings substantial gains in cross-dataset generalization (e.g., +2.95\% AUC on Celeb-DF when trained on FF++c23). This confirms that BAM effectively extracts and enhances forgery cues in the frequency domain that are more transferable.

\subsubsection{Variants of Frequency Modulation}
We compare different modulation strategies: (a) \textbf{Vanilla}: No frequency processing. (b) \textbf{FCAM}: Applies attention to each DCT coefficient (fine-grained). (c) \textbf{BAM}: Our band-wise attention. (d) \textbf{FGB/MGB}: Patch-based fine/mixed-grained BAM. Results in Table \ref{tab:BAM_or_vanilla} show that BAM strikes the best balance between performance and efficiency. FCAM, despite its finer granularity, does not outperform BAM and is computationally heavier. FGB and MGB introduce excessive complexity without commensurate gains, validating our design choice of operating on global frequency bands.

\subsubsection{Fine/Mix-Grained BAM (FGB/MGB)}
To simulate the inverse process of JPEG compression, we segment images (224×224) into 8×8 patches and apply DCT at this finer spatial granularity. Each patch is processed by an individual BAM module, resulting in 784 separate BAM modules. We also test Mixed-Grained BAM (MGB), which applies the same operations to patches of sizes 16×16 and 32×32. The resulting attention maps are then combined using a 1×1 convolution.
However, as shown in \cref{tab:BAM_or_vanilla}, both FGB and MGB yield suboptimal results compared to BAM and increase computational cost. This indicates that finer-grained or mixed-grained attention mechanisms may not be as effective as BAM in improving generalization within reasonable training budgets.

\begin{table}[tbp]
\caption{Ablation study on the impact of BAM. 
}
\centering
\setlength{\tabcolsep}{3pt}
\renewcommand{\arraystretch}{0.7}
\label{tab:eval_all}
\begin{tabular}{lcccc}
\hline
Methods & Training & \multicolumn{3}{c}{Testing} \\
& & Celeb & DFDC & FF++(c23 or c40)\\
\hline
\bf \color{red} ViR & FF++ & \bf  \color{red} 75.78 & \bf \color{red} 69.48 & \bf \color{red} 92.69\\
\bf BAM-Net & (c40) & \bf  78.74 & \bf  72.42 & \bf 93.19\\
\cline{1-5}
\bf \color{red} ViR & FF++ & \bf \color{red} 79.80 & \bf \color{red} 69.20 & \bf \color{red} 99.09\\
\bf BAM-Net & (c23) & \bf 81.85 & \bf 72.37 & \bf 99.32\\
\bottomrule
\end{tabular}
\end{table}

\begin{table}[tbp]
    \centering
    \caption{Ablation studies about different frequency attention methods and spatial segmentation granularities, on subsets of FF++c23 in terms of Acc(\%) and AUC(\%). 
    }
    \begin{tabular}{c|c|c|c|c|c}
        \hline
         Method & vanilla & FCAM & BAM & FGB & MGB  \\
         \hline
         Acc (\%) & 94.37 & 94.93 & \bf  95.03 & 93.40 & 94.10 \\
         AUC (\%) & 98.18 & 98.09 & \bf  98.35 & 97.50 & 98.12 \\
         \hline
    \end{tabular}
    \vspace{-1em}
    \label{tab:BAM_or_vanilla}
\end{table}

\section{Conclusions}
We presented BAM-Net, a framework for robust and generalizable face forgery detection. Its core is the Band Attention Modulation (BAM) mechanism, which dynamically reweights DCT frequency bands to enhance compression-resilient and transferable forensic cues. BAM is combined with the Vision Retentive (ViR) backbone, an efficient spatial encoder with distance-decayed attention.

Experiments on FF++, Celeb-DF, and DFDC show that BAM-Net achieves state-of-the-art performance and strong generalization in cross-dataset, cross-compression, and cross-manipulation settings. Ablation studies further validate the effectiveness of both BAM and ViR.

These results highlight the importance of adaptive frequency modulation for forgery detection. Future work will extend this framework to temporal modeling and other multimedia authenticity tasks.


\clearpage
\setcounter{section}{0}
\setcounter{subsection}{0}
\setcounter{table}{0}
\setcounter{figure}{0}
\setcounter{equation}{0}
\renewcommand{\theHsection}{supp.\arabic{section}}
\renewcommand{\theHsubsection}{supp.\arabic{section}.\arabic{subsection}}
\renewcommand{\theHtable}{supp.\arabic{table}}
\renewcommand{\theHfigure}{supp.\arabic{figure}}
\renewcommand{\theHequation}{supp.\arabic{equation}}
\twocolumn[{\centering\LARGE Supplementary Material for\\
\textit{Band-Attention Modulation Network for Robust Face Forgery Detection}\par\vspace{1.5em}}]

\maketitle

\section{Related Work}
\label{supp:related_work}

\subsection{Face Forgery Detection in the Spatial Domain}
Early deep learning-based face forgery detectors primarily operated in the spatial (RGB) domain, leveraging powerful convolutional neural networks (CNNs) to capture local texture inconsistencies or global semantic anomalies. A seminal work is FaceForensics++ \cite{supp:rossler2019faceforensics++}, which introduced a large-scale benchmark and demonstrated the effectiveness of fine-tuning CNNs like Xception on cropped face regions. Subsequent efforts have explored various architectures and training strategies to improve robustness and generalization. For instance, attention mechanisms have been incorporated to guide models toward more discriminative regions \cite{supp:dang2020detection}. Vision transformers have also been adapted for this task, modeling long-range dependencies to capture subtle artifacts \cite{supp:F2Trans}.

While these spatial-domain methods achieve high accuracy under intra-dataset settings, they often exhibit significant performance degradation when confronted with strong image compression (which obscures fine-grained textures) or novel forgery techniques not seen during training. This limitation highlights the need for complementary cues beyond pixel-level spatial patterns.

\subsection{Forgery Detection in the Frequency Domain}
The observation that manipulation operations often introduce distinctive spectral distortions has motivated a growing body of work exploring the frequency domain for forgery detection. These methods typically transform images using Discrete Cosine Transform (DCT) or Fourier Transform (FFT) and analyze the resulting spectrograms.

A key line of research applies fixed filters or hand-crafted operations to highlight high-frequency components, which are presumed to contain forgery traces. For example, high-pass filters are commonly used to suppress low-frequency content and amplify high-frequency noise patterns \cite{supp:xu2020learning}. Other works decompose the frequency spectrum into predefined bands (e.g., low, mid, high) and process them separately or fuse them with spatial features \cite{supp:F3Net}. This approach has been extended to video forgery detection through spatio-temporal frequency analysis \cite{supp:wang2023spatial} and to other domains like document forensics \cite{supp:qu2023towards}.

Despite their success, a fundamental limitation of most existing frequency-based methods is their \textbf{static nature}. They rely on pre-defined filters, fixed band separations, or heuristic operations that cannot adapt to the diverse and evolving spectral signatures of different forgery methods, compression levels, or image content. This lack of adaptability restricts their generalization capability and optimality.

\subsection{Adaptive Attention Mechanisms in Vision}
Attention mechanisms, which allow models to dynamically focus on informative parts of the input, have revolutionized computer vision. Channel Attention \cite{supp:hu2018squeeze} and Spatial Attention \cite{supp:woo2018cbam} modules enable networks to emphasize important features. The success of transformers further underscores the power of data-driven, content-adaptive weighting.

Our proposed \textbf{Band Attention Modulation (BAM)} draws inspiration from this paradigm but applies it uniquely in the frequency domain for forgery detection. Unlike static frequency filters, BAM introduces a \textit{learnable, lightweight network} to dynamically reweight different frequency bands based on the input content. This represents a shift from \textit{filtering} to \textit{modulation}, enabling the model to automatically discover and emphasize the most discriminative spectral components for a given sample—a novel contribution that bridges the gap between adaptive attention and frequency-domain analysis for forensic tasks.

\section{Experimental Details}
\label{supp:exp_details}

\subsection{Datasets}
We use the following datasets for evaluation:
\begin{itemize}
    \item \textbf{FaceForensics++ (FF++)} \cite{supp:rossler2019faceforensics++}: Contains 1,000 real and 4,000 fake videos generated by four methods: Face2Face (FF), FaceSwap (FS), DeepFakes (DF), and NeuralTextures (NT). We use both High Quality (c23) and Low Quality (c40) versions. Following the common split \cite{supp:F2Trans}, we use 720 videos for training, 140 for validation, and 140 for testing per 1,000-video set. For testing, we extract $\sim$400 frames per video; for training, we use only 20 frames per video to simulate data efficiency.
    \item \textbf{Celeb-DF} \cite{supp:li2020celeb}: Contains 590 real and 5,639 high-quality fake videos. We use the same frame sampling strategy as FF++.
    \item \textbf{DFDC} \cite{supp:DFDC}: A large-scale dataset with 128,154 videos. We follow the same frame extraction protocol.
    \item \textbf{GenImage} \cite{supp:zhu2024genimage}: Used for non-face AI-generated image detection. It contains 1.33M real images from ImageNet and 1.35M fake images generated by models including Stable Diffusion (SD), ADM, GLIDE, BigGAN, Midjourney, and VQDM.
\end{itemize}

\subsection{Evaluation Metrics}
We report Area Under the ROC Curve (AUC) and Accuracy (Acc) at two levels:
\begin{itemize}
    \item \textbf{Image-Level (IL)}: Each frame is predicted independently.
    \item \textbf{Image-based Video-Level (IVL)}: The score for a video is the average of its frame predictions.
\end{itemize}

\subsection{Implementation Details}
Faces are detected and aligned using DLIB \cite{supp:sagonas2016300}. Input images are resized to $224\times224$. We apply random erasing for augmentation. The model is trained with the Adam optimizer ($\beta_1=0.9$, $\beta_2=0.99$) for 20 epochs with early stopping. The initial learning rate is $10^{-4}$.

\section{Additional Experiments and Analysis}
\label{supp:additional}

\subsection{Sensitivity to Training Data Size}
To evaluate data efficiency, we compare two training subsets from FF++c23:
\begin{itemize}
    \item \textbf{TRAIN-ORI}: 20 frames per video (standard setting).
    \item \textbf{TRAIN-SUB}: 4 randomly sampled frames per video.
\end{itemize}
Results in Table~\ref{tab:set_size} show that TRAIN-SUB achieves performance close to TRAIN-ORI, demonstrating our model's robustness even with limited data.

\begin{table}[htbp]
    \centering
    \caption{Performance with different training set sizes on FF++c23.}
    \setlength{\tabcolsep}{5pt}
    \begin{tabular}{l|cc|cc}
        \hline
        Size & Acc(IL) & AUC(IL) & Acc(IVL) & AUC(IVL) \\
        \hline
        SUB  & 96.11 & 98.84 & 98.14 & 99.66 \\
        ORI  & 97.02 & 99.32 & 98.00 & 99.81 \\
        \hline
    \end{tabular}
    \label{tab:set_size}
\end{table}

\begin{table*}[t]
\caption{Cross-generator detection results on GenImage. Models are trained on SD v1.4, and accuracy (\%) is reported. \textbf{Bold} highlights the best performance, and \underline{underline} indicates the second best.}
\centering
\label{tab:GenImage}
\begin{tabular}{@{}lcccccccc|c@{}}
\toprule
Method & SD v1.4 & SD v1.5 & Wukong & ADM & BigGAN & GLIDE & Midjourney & VQDM & Average\\
\midrule
$F^3$-Net
& \bf 99.9& \bf 99.9& \bf 99.9& 49.9 & 49.9& 50.0 & 50.1 & 49.9 & 68.7\\
GramNet
&99.2&99.1&98.9&50.3& \underline{51.7}&54.6&54.2&50.8&69.9\\
PatchFron
&99.7&99.4&96.7&51.0&51.6&54.1&66.2& \bf 54.1&71.6\\
SIA
&\underline{99.8}&\underline{99.6}&98.0&52.0&50.3&52.3&\underline{71.6}&51.4&71.9\\
LGrad
&90.3&64.7&63.9& \underline{52.2}& \bf 69.9&\underline{62.7} &56.5& 52.0&64.0\\
Ours
& \underline{99.8}& \underline{99.6}& \underline{99.2}& \bf 62.4&51.4& \bf 68.0& \bf 73.6&\underline{52.8}& \bf 75.8\\
\bottomrule
\end{tabular}
\end{table*}

\begin{table}[htbp]
    \centering
    \caption{Ablation of BAM design choices on FF++c23.}
    \begin{tabular}{l|c|c|c}
        \hline
        Method & BAM (Ours) & Softmax BAM & Concatenate BAM \\
        \hline
        Acc (\%) & \bf 95.03 & 94.43 & 94.80 \\
        AUC (\%) & \bf 98.35 & 97.94 & 98.17 \\
        \hline
    \end{tabular}
    \label{tab:diff_BAM}
\end{table}

\begin{table}[htbp]
\caption{Efficiency comparison (parameters, FLOPs, throughput).}
\centering
\renewcommand{\arraystretch}{0.7}
\label{tab:params}
\begin{tabular}{@{}lccc@{}}
\toprule
Model & Params & FLOPs & Throughput (imgs/s) \\
\midrule
Swin-B & 88M & 15.4G & 756 \\
EfficientNet-B6 & 43M & 19.0G & 172 \\
MaxViT-B & 120M & 23.4G & 306 \\
F$^2$Trans-B & 128M & 21.8G & -- \\
\textbf{BAM-Net (Ours)} & \bf 96M & \bf 18.2G & \bf 326 \\
\bottomrule
\end{tabular}
\end{table}

\subsection{Non-face Pictures Evaluation}

We test BAM-Net's ability to detect AI-generated images beyond faces using the GenImage benchmark. Following the standard protocol, we train on the SD v1.4 subset and test on all eight subsets. Results in Table~\ref{tab:GenImage} show that BAM-Net achieves the highest average accuracy (75.8\%), outperforming specialized detectors. It performs robustly across both diffusion-based (SD, ADM) and GAN-based (BigGAN, GLIDE) methods, demonstrating its generalizable frequency-based reasoning.

\subsection{Design Choices in BAM}
We ablate two key design decisions in BAM:
\begin{itemize}
    \item \textbf{Pooling fusion}: Adding vs. concatenating GMP and GAP features.
    \item \textbf{Weight normalization}: Sigmoid vs. Softmax.
\end{itemize}
As shown in Table~\ref{tab:diff_BAM}, simply adding the pooled features and using Sigmoid yields the best performance, leading to our final simple design.

\subsection{Efficiency Analysis}
Table~\ref{tab:params} compares the computational cost of BAM-Net with other backbones. BAM-Net achieves a favorable trade-off: higher accuracy than larger models (e.g., MaxViT-B, F$^2$Trans-B) with fewer parameters and competitive FLOPs.

\section{Visualizations}
\label{supp:visual}

\subsection{BAM Modulation Effects}
\begin{figure}[htbp]
  \centering
  \includegraphics[width=0.45\textwidth]{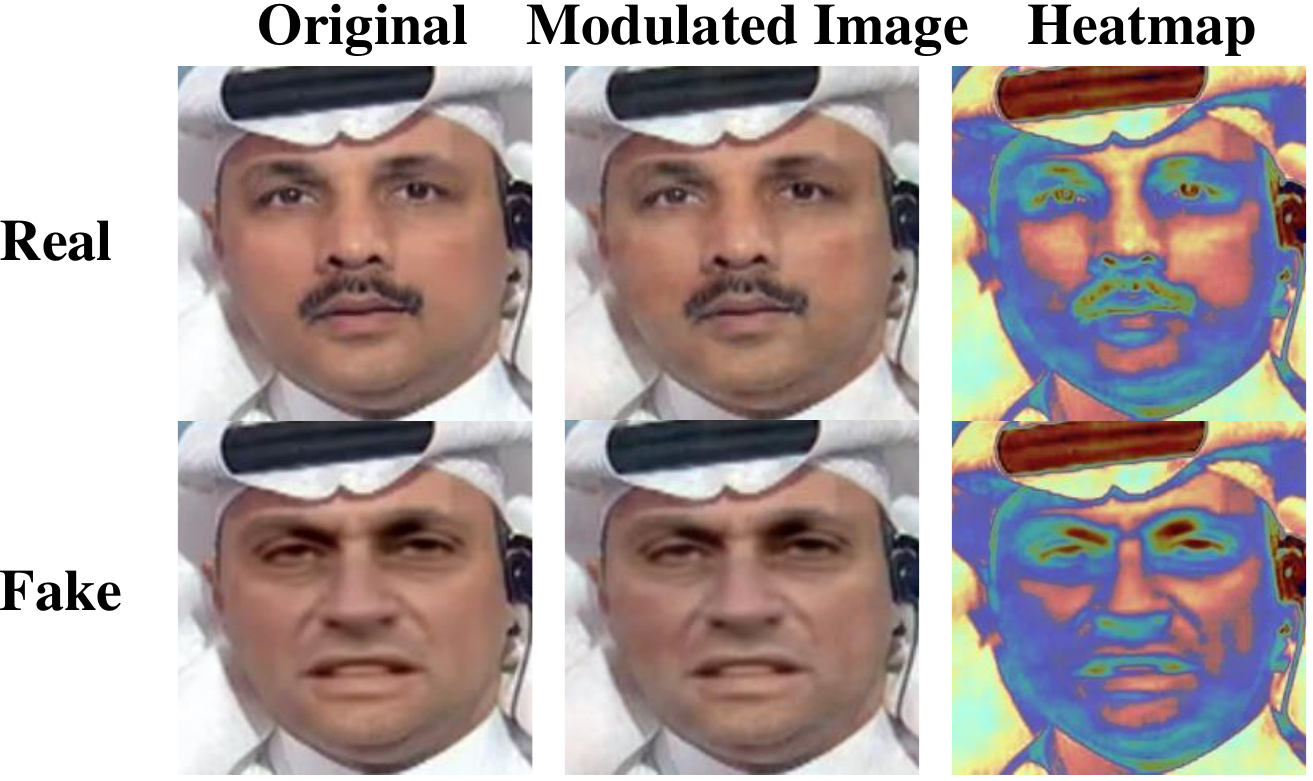}
  \caption{Visualization of BAM’s effect. Left: original images. Middle: modulated images after BAM and iDCT. Right: spatial attention maps derived from modulation. Differences between real and fake samples are visible in the attention patterns.}
  \label{fig:visual_BAM}
\end{figure}

Figure~\ref{fig:visual_BAM} visualizes the output of the BAM module. The modulated images (after iDCT) show subtle enhancements in high-frequency regions compared to the originals. The corresponding spatial attention maps highlight different patterns for real vs. fake images, indicating that BAM learns to accentuate forgery-sensitive regions.

\subsection{Grad-CAM and Grad-CAM++}
\begin{figure*}[htbp]
  \centering
  \includegraphics[width=0.84\textwidth]{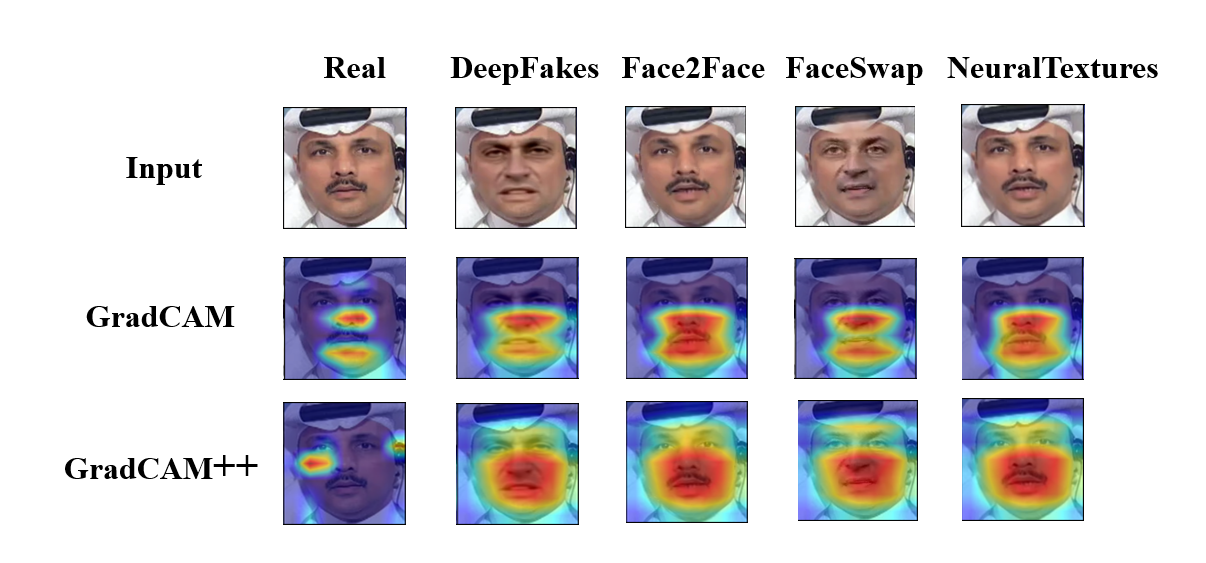}
  \includegraphics[width=0.82\textwidth]{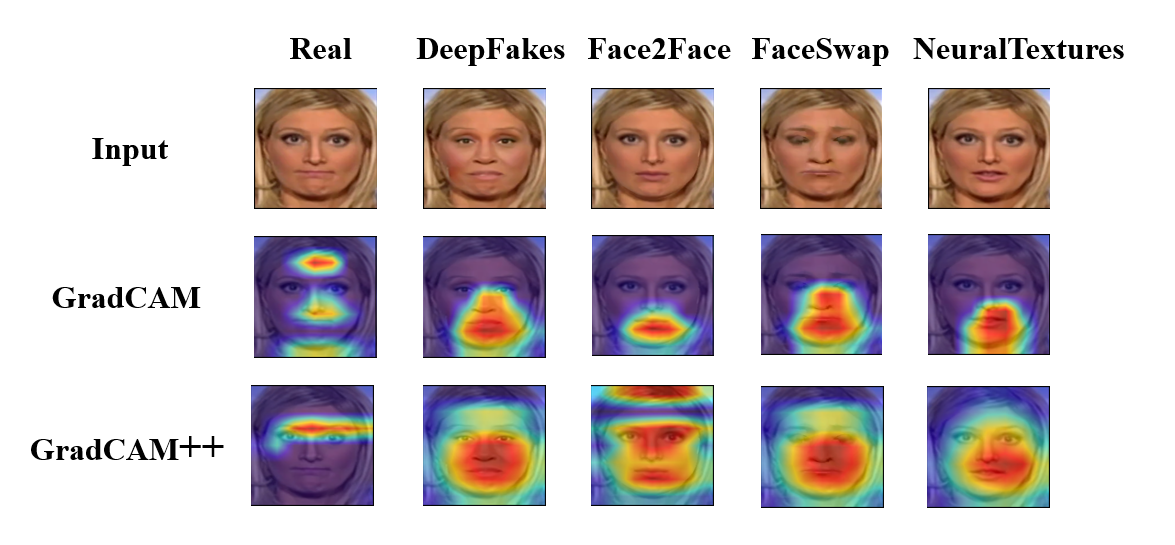}
  \includegraphics[width=0.8\textwidth]{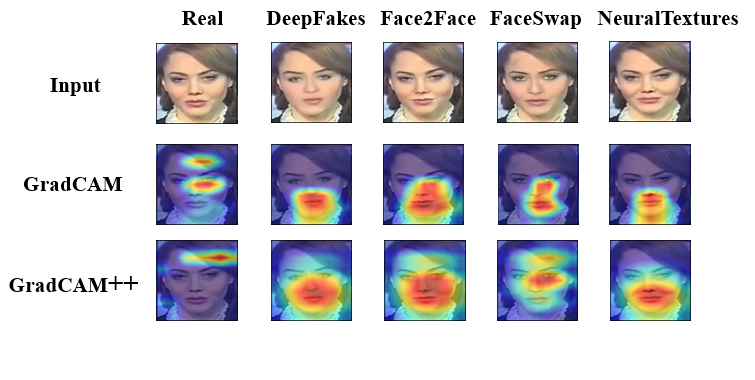}
  \caption{Grad-CAM and Grad-CAM++ visualizations for BAM-Net on three sample IDs. The model shows more localized, high-intensity attention for fake faces compared to real ones.}
  \label{fig:GradCAM}
\end{figure*}

We apply Grad-CAM \cite{supp:selvaraju2017grad} and Grad-CAM++ \cite{supp:chattopadhay2018grad} to visualize where BAM-Net focuses when making predictions. As shown in Figure~\ref{fig:GradCAM}, for fake images, the model consistently attends to specific facial regions (e.g., eyes, mouth boundaries) where artifacts are common, whereas for real images, the attention is more diffuse.

\section*{List of Compared Methods}
For completeness, we list all methods compared in our experiments: Xception \cite{supp:rossler2019faceforensics++}, SPSL \cite{supp:SPSL}, $F^3$-Net \cite{supp:F3Net}, M2TR \cite{supp:M2TR}, LiSiam \cite{supp:lisiam}, SIA \cite{supp:sun2022information}, DisGRL \cite{supp:shi2023discrepancy}, $F^2$Trans-B \cite{supp:F2Trans}, Two-branch \cite{supp:two_branch}, MADD \cite{supp:MADD}, Face X-ray \cite{supp:face_x_ray}, LTW \cite{supp:LTW}, FTCN \cite{supp:FTCN}, EfficientNet \cite{supp:efficientnet}, ForensicTransfer \cite{supp:forensictransfer}, Multi-task \cite{supp:multi_task}, MLDG \cite{supp:MLDG}, DCL \cite{supp:DCL}, LGrad \cite{supp:LGrad}, DDIN \cite{supp:DDIN}, DPBoT \cite{supp:DPBoT}, DeepFake-Adapter \cite{supp:DeepFake-Adapter}, FDPNet \cite{supp:li2025fdpnet}, MTSD \cite{supp:MTSD}.

\begin{thebibliography}{10}

\bibitem{rossler2019faceforensics++}
Andreas Rossler, Davide Cozzolino, Luisa Verdoliva, Christian Riess, Justus Thies, and Matthias Nie{\ss}ner,
\newblock ``Faceforensics++: Learning to detect manipulated facial images,''
\newblock in {\em CVPR}, 2019, pp. 1--11.

\bibitem{F2Trans}
Changtao Miao, Zichang Tan, Qi~Chu, Huan Liu, Honggang Hu, and Nenghai Yu,
\newblock ``F 2 trans: High-frequency fine-grained transformer for face forgery detection,''
\newblock {\em IEEE TIFS}, vol. 18, pp. 1039--1051, 2023.

\bibitem{duan2025dual}
Junxian Duan, Siyu Liu, Yiming Hao, Huaibo Huang, and Ran He,
\newblock ``Dual frequency-guided spatiotemporal feature learning for face forgery detection,''
\newblock {\em IEEE Transactions on Biometrics, Behavior, and Identity Science}, 2025.

\bibitem{xu2020learning}
Kai Xu, Minghai Qin, Fei Sun, Yuhao Wang, Yen-Kuang Chen, and Fengbo Ren,
\newblock ``Learning in the frequency domain,''
\newblock in {\em CVPR}, 2020, pp. 1740--1749.

\bibitem{F3Net}
Yuyang Qian, Guojun Yin, Lu~Sheng, Zixuan Chen, and Jing Shao,
\newblock ``Thinking in frequency: Face forgery detection by mining frequency-aware clues,''
\newblock in {\em ECCV}. Springer, 2020, pp. 86--103.

\bibitem{RetNet}
Yutao Sun, Li~Dong, Shaohan Huang, Shuming Ma, Yuqing Xia, Jilong Xue, Jianyong Wang, and Furu Wei,
\newblock ``Retentive network: A successor to transformer for large language models,''
\newblock {\em arXiv preprint arXiv:2307.08621}, 2023.

\bibitem{su2021roformer}
Jianlin Su, Yu~Lu, Shengfeng Pan, Ahmed Murtadha, Bo~Wen, and Yunfeng Liu,
\newblock ``Roformer: Enhanced transformer with rotary position embedding,'' 2023.

\bibitem{ViT}
Alexey Dosovitskiy,
\newblock ``An image is worth 16x16 words: Transformers for image recognition at scale,''
\newblock {\em arXiv preprint arXiv:2010.11929}, 2020.

\bibitem{li2020celeb}
Yuezun Li, Xin Yang, Pu~Sun, Honggang Qi, and Siwei Lyu,
\newblock ``Celeb-df: A large-scale challenging dataset for deepfake forensics,''
\newblock in {\em CVPR}, 2020, pp. 3207--3216.

\bibitem{DFDC}
Brian Dolhansky, Joanna Bitton, Ben Pflaum, Jikuo Lu, Russ Howes, Menglin Wang, and Cristian~Canton Ferrer,
\newblock ``The deepfake detection challenge (dfdc) dataset,''
\newblock {\em arXiv preprint arXiv:2006.07397}, 2020.

\bibitem{FTCN}
Yinglin Zheng, Jianmin Bao, Dong Chen, Ming Zeng, and Fang Wen,
\newblock ``Exploring temporal coherence for more general video face forgery detection,''
\newblock in {\em ICCV}, 2021, pp. 15044--15054.

\end{thebibliography}

\begin{thebibliography}{10}

\bibitem{supp:rossler2019faceforensics++}
Andreas Rossler, Davide Cozzolino, Luisa Verdoliva, Christian Riess, Justus Thies, and Matthias Nie{\ss}ner,
\newblock ``Faceforensics++: Learning to detect manipulated facial images,''
\newblock in {\em CVPR}, 2019, pp. 1--11.

\bibitem{supp:dang2020detection}
Hao Dang, Feng Liu, Joel Stehouwer, Xiaoming Liu, and Anil~K Jain,
\newblock ``On the detection of digital face manipulation,''
\newblock in {\em CVPR}, 2020, pp. 5781--5790.

\bibitem{supp:F2Trans}
Changtao Miao, Zichang Tan, Qi~Chu, Huan Liu, Honggang Hu, and Nenghai Yu,
\newblock ``F 2 trans: High-frequency fine-grained transformer for face forgery detection,''
\newblock {\em IEEE TIFS}, vol. 18, pp. 1039--1051, 2023.

\bibitem{supp:xu2020learning}
Kai Xu, Minghai Qin, Fei Sun, Yuhao Wang, Yen-Kuang Chen, and Fengbo Ren,
\newblock ``Learning in the frequency domain,''
\newblock in {\em CVPR}, 2020, pp. 1740--1749.

\bibitem{supp:F3Net}
Yuyang Qian, Guojun Yin, Lu~Sheng, Zixuan Chen, and Jing Shao,
\newblock ``Thinking in frequency: Face forgery detection by mining frequency-aware clues,''
\newblock in {\em ECCV}. Springer, 2020, pp. 86--103.

\bibitem{supp:wang2023spatial}
Yukai Wang, Chunlei Peng, Decheng Liu, Nannan Wang, and Xinbo Gao,
\newblock ``Spatial-temporal frequency forgery clue for video forgery detection in vis and nir scenario,''
\newblock {\em IEEE TCSVT}, vol. 33, no. 12, pp. 7943--7956, 2023.

\bibitem{supp:qu2023towards}
Chenfan Qu, Chongyu Liu, Yuliang Liu, Xinhong Chen, Dezhi Peng, Fengjun Guo, and Lianwen Jin,
\newblock ``Towards robust tampered text detection in document image: New dataset and new solution,''
\newblock in {\em CVPR}, 2023, pp. 5937--5946.

\bibitem{supp:hu2018squeeze}
Jie Hu, Li~Shen, and Gang Sun,
\newblock ``Squeeze-and-excitation networks,''
\newblock in {\em Proceedings of the IEEE conference on computer vision and pattern recognition}, 2018, pp. 7132--7141.

\bibitem{supp:woo2018cbam}
Sanghyun Woo, Jongchan Park, Joon-Young Lee, and In~So Kweon,
\newblock ``Cbam: Convolutional block attention module,''
\newblock in {\em ECCV}, 2018, pp. 3--19.

\bibitem{supp:li2020celeb}
Yuezun Li, Xin Yang, Pu~Sun, Honggang Qi, and Siwei Lyu,
\newblock ``Celeb-df: A large-scale challenging dataset for deepfake forensics,''
\newblock in {\em CVPR}, 2020, pp. 3207--3216.

\bibitem{supp:DFDC}
Brian Dolhansky, Joanna Bitton, Ben Pflaum, Jikuo Lu, Russ Howes, Menglin Wang, and Cristian~Canton Ferrer,
\newblock ``The deepfake detection challenge (dfdc) dataset,''
\newblock {\em arXiv preprint arXiv:2006.07397}, 2020.

\bibitem{supp:zhu2024genimage}
Mingjian Zhu, Hanting Chen, Qiangyu Yan, Xudong Huang, Guanyu Lin, Wei Li, Zhijun Tu, Hailin Hu, Jie Hu, and Yunhe Wang,
\newblock ``Genimage: A million-scale benchmark for detecting ai-generated image,''
\newblock {\em NeurIPS}, vol. 36, 2024.

\bibitem{supp:sagonas2016300}
Christos Sagonas, Epameinondas Antonakos, Georgios Tzimiropoulos, Stefanos Zafeiriou, and Maja Pantic,
\newblock ``300 faces in-the-wild challenge: Database and results,''
\newblock {\em Image Vision Comput.}, vol. 47, pp. 3--18, 2016.

\bibitem{supp:selvaraju2017grad}
Ramprasaath~R Selvaraju, Michael Cogswell, Abhishek Das, Ramakrishna Vedantam, Devi Parikh, and Dhruv Batra,
\newblock ``Grad-cam: Visual explanations from deep networks via gradient-based localization,''
\newblock in {\em Proceedings of the IEEE international conference on computer vision}, 2017, pp. 618--626.

\bibitem{supp:chattopadhay2018grad}
Aditya Chattopadhay, Anirban Sarkar, Prantik Howlader, and Vineeth~N Balasubramanian,
\newblock ``Grad-cam++: Generalized gradient-based visual explanations for deep convolutional networks,''
\newblock in {\em 2018 IEEE winter conference on applications of computer vision (WACV)}. IEEE, 2018, pp. 839--847.

\bibitem{supp:SPSL}
Honggu Liu, Xiaodan Li, Wenbo Zhou, Yuefeng Chen, Yuan He, Hui Xue, Weiming Zhang, and Nenghai Yu,
\newblock ``Spatial-phase shallow learning: rethinking face forgery detection in frequency domain,''
\newblock in {\em CVPR}, 2021, pp. 772--781.

\bibitem{supp:M2TR}
Junke Wang, Zuxuan Wu, Wenhao Ouyang, Xintong Han, Jingjing Chen, Yu-Gang Jiang, and Ser-Nam Li,
\newblock ``M2tr: Multi-modal multi-scale transformers for deepfake detection,''
\newblock in {\em ICMR}, 2022, pp. 615--623.

\bibitem{supp:lisiam}
Jian Wang, Yunlian Sun, and Jinhui Tang,
\newblock ``Lisiam: Localization invariance siamese network for deepfake detection,''
\newblock {\em IEEE TIFS}, vol. 17, pp. 2425--2436, 2022.

\bibitem{supp:sun2022information}
Ke~Sun, Hong Liu, Taiping Yao, Xiaoshuai Sun, Shen Chen, Shouhong Ding, and Rongrong Ji,
\newblock ``An information theoretic approach for attention-driven face forgery detection,''
\newblock in {\em ECCV}. Springer, 2022, pp. 111--127.

\bibitem{supp:shi2023discrepancy}
Zenan Shi, Haipeng Chen, Long Chen, and Dong Zhang,
\newblock ``Discrepancy-guided reconstruction learning for image forgery detection,''
\newblock {\em arXiv preprint arXiv:2304.13349}, 2023.

\bibitem{supp:two_branch}
Iacopo Masi, Aditya Killekar, Royston~Marian Mascarenhas, Shenoy~Pratik Gurudatt, and Wael AbdAlmageed,
\newblock ``Two-branch recurrent network for isolating deepfakes in videos,''
\newblock in {\em Computer Vision--ECCV 2020: 16th European Conference, Glasgow, UK, August 23--28, 2020, Proceedings, Part VII 16}. Springer, 2020, pp. 667--684.

\bibitem{supp:MADD}
Hanqing Zhao, Wenbo Zhou, Dongdong Chen, Tianyi Wei, Weiming Zhang, and Nenghai Yu,
\newblock ``Multi-attentional deepfake detection,''
\newblock in {\em CVPR}, 2021, pp. 2185--2194.

\bibitem{supp:face_x_ray}
Lingzhi Li, Jianmin Bao, Ting Zhang, Hao Yang, Dong Chen, Fang Wen, and Baining Guo,
\newblock ``Face x-ray for more general face forgery detection,''
\newblock in {\em CVPR}, 2020, pp. 5001--5010.

\bibitem{supp:LTW}
Ke~Sun, Hong Liu, Qixiang Ye, Yue Gao, Jianzhuang Liu, Ling Shao, and Rongrong Ji,
\newblock ``Domain general face forgery detection by learning to weight,''
\newblock in {\em AAAI}, 2021, vol.~35, pp. 2638--2646.

\bibitem{supp:FTCN}
Yinglin Zheng, Jianmin Bao, Dong Chen, Ming Zeng, and Fang Wen,
\newblock ``Exploring temporal coherence for more general video face forgery detection,''
\newblock in {\em ICCV}, 2021, pp. 15044--15054.

\bibitem{supp:efficientnet}
Mingxing Tan and Quoc Le,
\newblock ``Efficientnet: Rethinking model scaling for convolutional neural networks,''
\newblock in {\em ICML}. PMLR, 2019, pp. 6105--6114.

\bibitem{supp:forensictransfer}
Davide Cozzolino, Justus Thies, Andreas R{\"o}ssler, Christian Riess, Matthias Nie{\ss}ner, and Luisa Verdoliva,
\newblock ``Forensictransfer: Weakly-supervised domain adaptation for forgery detection,''
\newblock {\em arXiv preprint arXiv:1812.02510}, 2018.

\bibitem{supp:multi_task}
Huy~H Nguyen, Fuming Fang, Junichi Yamagishi, and Isao Echizen,
\newblock ``Multi-task learning for detecting and segmenting manipulated facial images and videos,''
\newblock in {\em BTAS}. IEEE, 2019, pp. 1--8.

\bibitem{supp:MLDG}
Da~Li, Yongxin Yang, Yi-Zhe Song, and Timothy Hospedales,
\newblock ``Learning to generalize: Meta-learning for domain generalization,''
\newblock in {\em AAAI}, 2018, vol.~32.

\bibitem{supp:DCL}
Ke~Sun, Taiping Yao, Shen Chen, Shouhong Ding, Jilin Li, and Rongrong Ji,
\newblock ``Dual contrastive learning for general face forgery detection,''
\newblock in {\em AAAI}, 2022, vol.~36, pp. 2316--2324.

\bibitem{supp:LGrad}
Chuangchuang Tan, Yao Zhao, Shikui Wei, Guanghua Gu, and Yunchao Wei,
\newblock ``Learning on gradients: Generalized artifacts representation for gan-generated images detection,''
\newblock in {\em CVPR}, 2023, pp. 12105--12114.

\bibitem{supp:DDIN}
Shenyuan Huang, Huaibo Huang, Zi~Wang, Nan Xu, Aihua Zheng, and Ran He,
\newblock ``Uncertainty-guided test-time training for face forgery detection,''
\newblock in {\em ACPR}. Springer, 2023, pp. 258--272.

\bibitem{supp:DPBoT}
Kaihan Lin, Jujian Lv, Huimin Zhao, Jinchang Ren, Shudong Li, Weihong Han, Jiawen Li, and Zhaoquan Gu,
\newblock ``Dual-branch pre-activation bottleneck transformer for face forgery detection,''
\newblock {\em Cognitive Computation}, vol. 17, no. 3, pp. 118, 2025.

\bibitem{supp:DeepFake-Adapter}
Rui Shao, Tianxing Wu, Liqiang Nie, and Ziwei Liu,
\newblock ``Deepfake-adapter: Dual-level adapter for deepfake detection,''
\newblock {\em International Journal of Computer Vision}, vol. 133, no. 6, pp. 3613--3628, 2025.

\bibitem{supp:li2025fdpnet}
Boyang Li, Huihuang Zhao, and Leyi Li,
\newblock ``Fdpnet: deep forgery detection by leveraging multi-scale self-forgery images generating,''
\newblock {\em The Journal of Supercomputing}, vol. 81, no. 6, pp. 1--24, 2025.

\bibitem{supp:MTSD}
Dexu Zhu, Jie Cao, Jiangnan Shao, Zhida Zhang, Junxian Duan, and Ran He,
\newblock ``Mtsd: Simple yet effective self-distillation for generalizable deepfake detection,''
\newblock in {\em 2025 IEEE International Conference on Multimedia and Expo (ICME)}, 2025, pp. 1--6.

\end{thebibliography}
\end{document}